\documentclass[a4paper,fleqn]{cas-dc}

\usepackage[authoryear,longnamesfirst]{natbib}
\usepackage{url}
\def\tsc#1{\csdef{#1}{\textsc{\lowercase{#1}}\xspace}}
\tsc{WGM}
\tsc{QE}

\begin{document}
\let\WriteBookmarks\relax
\def\floatpagepagefraction{1}
\def\textpagefraction{.001}

\shorttitle{HGraphormer}    

\shortauthors{S. Qu, W. Wang et al.}  

\title [mode = title]{Hypergraph Node Representation Learning with One-Stage Message Passing}  

\author{Shilin Qu}
\ead{shilin.qu@monash.edu}
\fnmark[1]
\fntext[1]{Faculty of Information Technology, Monash University, Melbourne, VIC, 3800, Australia}
\fntext[2]{AI Division, Westlake University, Hangzhou, Zhejiang, 310024, P.R. China}

\author{Weiqing Wang}
\ead{teresa.wang@monash.edu}
\fnmark[1]
\author{Yuan-Fang Li}
\ead{YuanFang.Li@monash.edu}
\fnmark[1]

\author{Xin Zhou}
\ead{xin.zhou@monash.edu}
\fnmark[1]

\author{Fajie Yuan}
\ead{yuanfajie@westlake.edu.cn}
\fnmark[2]

\cortext[1]{Corresponding author: Shilin Qu, Weiqing Wang}

\begin{abstract}
Hypergraphs as an expressive and general structure have attracted considerable attention from various research domains. Most existing hypergraph node representation learning techniques are based on graph neural networks, and thus adopt the two-stage message passing paradigm (i.e.\ node $\rightarrow$ hyperedge $\rightarrow$ node). This paradigm only focuses on local information propagation and does not effectively take into account global information, resulting in less optimal representations. Our theoretical analysis of representative two-stage message passing methods shows that, mathematically, they model different ways of local message passing through hyperedges, and can be unified into one-stage message passing (i.e.\ node $\rightarrow$ node). However, they still only model local information. Motivated by this theoretical analysis, we propose a novel one-stage message passing paradigm to model both global and local information propagation for hypergraphs. We integrate this paradigm into HGraphormer, a Transformer-based framework for hypergraph node representation learning. HGraphormer injects the hypergraph structure information (local information) into Transformers (global information) by combining the attention matrix and hypergraph Laplacian. Extensive experiments demonstrate that HGraphormer outperforms recent hypergraph learning methods on five representative benchmark datasets on the semi-supervised hypernode classification task, setting new state-of-the-art performance, with accuracy improvements between 2.52\% and 6.70\%. 
Our code and datasets are available\footnote{\url{https://anonymous.4open.science/r/HGraphormer-DF3F}}.
\end{abstract}

\begin{keywords}
Hypergraph \sep Graph \sep Transformer\sep Node Representation
\end{keywords}

\maketitle

\section{Introduction}\label{sec:intro}
\begin{figure}[h]
    \centering
    \includegraphics[width=240pt,height=130pt]{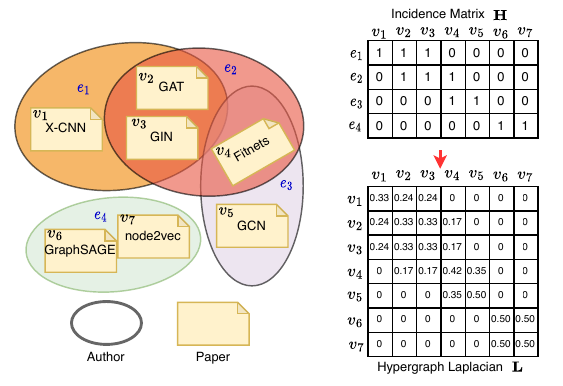} 
    \caption{(left) Paper-Authorship Hypergraph. The nodes represent papers and multiple papers written by one author form a hyperedge. (right) The incidence matrix $\mathbf{H}$ and hypergraph Laplacian $\mathbf{L}$ of Paper-Authorship Hypergraph. $e_1,...,e_4$ and $v_1,...,v_7$ are hyperedges and nodes, respectively.}
    \label{fig:coauthor}
\end{figure}

Compared with regular graphs, hypergraphs~\cite{hypergcn2019,HGNN2019,yoon2020howmuch,amburg2020clustergraph,UniGNN2021,liu2021diffsemi} is a more flexible and general data structure that allows an edge to join any number of nodes, which can naturally capture high-order relationships among multiple nodes. Take the paper-authorship hypergraph in Figure~\ref{fig:coauthor}(left) as an example, nodes ($v_1, v_2, \ldots$) represent papers, and papers written by the same author form a hyperedge ($e_1, e_2, \ldots$). Hypergraph has attracted much attention in a wide range of fields, including computer vision~\cite{HGNNPlus2022,ijcai2020p109}, natural sciences~\cite{PhysRevA2020}, social networks~\cite{yu2021selfsocialrec,sun2021multisocial}, 
and recommendation systems~\cite{reshgnn2021,xue2021mbn,cheng2022ihgnn}.

\begin{figure}[h]
    \centering
    \includegraphics[height=80pt]{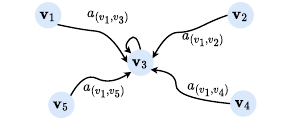} 
    \caption{One-stage message passing for GNNs. $\mathbf{v}_i$ is the features of node $v_i$. $a_{(v_i,v_j)}$ is the weight between node $v_i$ and node $v_j$.}
    \label{fig-mpass3}
\end{figure}

\begin{figure}[h]
    \centering
    \includegraphics[height=80pt]{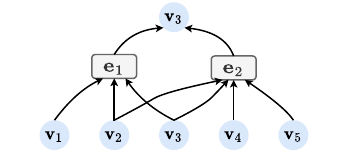} 
    \caption{Two-stage message passing widely used in HGNNs. $\mathbf{v}_i$ and $\mathbf{e}_j$ are the features of node $v_i$ and hyperedge $e_j$ respectively.}
    \label{fig-mpass}
\end{figure}

Message passing is the foundation of graph neural networks (GNNs)~\cite{GCN2017,GAT2018,GTNs2019,Graphormer2021} and hypergraph neural networks (HGNNs)~\cite{HGNNPlus2022,xue2021mbn,cheng2022ihgnn,ijcai2020p109,hypergcn2019,HGNN2019,UniGNN2021}. The ultimate goal of this mechanism is to learn good node 
representations for downstream tasks. (Hyper)edges serve as a bridge for information transmission, transfer and aggregation. Conceptually, GNNs utilises one-stage message passing, directly from nodes to nodes (i.e., node$\rightarrow$node), as illustrated in Figure~\ref{fig-mpass3}. In contrast, HGNNs utilise two-stage message passing~\cite{choe2022midas,bera2022densestsub,lee2021how}, firstly from nodes to hyperedges, then from hyperedges to nodes (i.e., node$\rightarrow$hyperedge$\rightarrow$node), as illustrated in Figure~\ref{fig-mpass}. 
Such two-stage message passing is intuitively motivated by the nature of hypergraphs, that a node can be connected by multiple nodes through a hyperedge. 
Compared with one-stage message passing, two-stage message passing transfers information between nodes indirectly through hyperedges.

As one of the most popular representation learning frameworks, Transformer~\cite{Vaswani2017} has demonstrated state-of-the-art performance in many domains, such as natural language processing~\cite{hou2022multigranularity,yu2021crosslingual,fei2021latent,zhao2020syncretic,zhao2020condition}, audio~\cite{haghani2018audio,chung2020generative}, computer vision~\cite{arnab2021vivit,khan2021Transformers,liu2021swin},
and graphs~\cite{Graphormer2021,dwivedi2020generalization,hu2020heterogeneous,zhang2020graph}.
By carefully designing position embeddings and the attention matrix, graph structures can be naturally injected into Transformer~\cite{hu2020heterogeneous,yang2020html,nguyen2022universalgraph}. For instance, Graphormer~\cite{Graphormer2021} integrates attention and adjacency matrix to control the message passing between nodes and achieves state-of-the-art results on multiple real-world datasets (e.g., MolHIV, MolPCBA, ZINC). 

Following the two-stages message passing paradigm, some recent works on hypergraph representation learning employ Transformer as the encoder module at each stage of message passing~\cite{georgiev2022heat,chien2022you}. More specifically, they first use Transformer to encode nodes to obtain the hyperedge representations, then use Transformer to encode hyperedges to generate node representations.  

A fundamental deficiency of the two-stage message passing paradigm is its inability to effectively incorporate global information. Global information means the directed interaction (latent correlation) between nodes no matter whether there is a path between them. However, in two-stage message passing, interaction between nodes only happen via shared hyperedge and there must be at least one path between these two nodes.
In other words, it only enables local information passing but lacks global information. For example, in Figure~\ref{fig:coauthor}(left), node $v_2$ must undergo four message passing steps to interact with node $v_5$, through the path $v_2 \to e_2 \to v_4; v_4 \to e_3 \to v_5$. Furthermore, there is no path to connect node $v_2$ and node $v_6$. No matter how many times the message is propagated, no information interaction will happen between node $v_2$ and node $v_6$. But in practice, $v_2$ is an improvement work on $v_5$ and $v_6$, which means node $v_2$, $v_5$ and $v_6$ are very closely related.  In other words, such latent semantic correlation (global information) between nodes is also essential for modeling node representation.

In this paper, we conduct a theoretical analysis on the two-stage messaging passing mechanism and derive that it is a special case of one-stage message passing (Sec.\ref{theory_analysis}). In addition, we also observe that the Transformer architecture is amenable to one-stage message passing. 

Based on these findings, we propose {HGraphormer}, a novel framework that deeply integrates \textbf{H}yper\textbf{Graph} and Transf\textbf{ormer} that is able to directly and fully exploit Transformer' strong representation powers (Sec.~\ref{sec:hgraphormer}) instead of simply using Transformer as encoders like existing Transformer+hypergraphs works~\cite{georgiev2022heat,chien2022you}.  Distinct from all existing hypergraph node representation learning techniques, HGraphormer employs the one-stage message passing paradigm, which is able to incorporate both local and global interactions between nodes, resulting in better node representations. This is achieved by injecting a hypergraph's structure information into Transformer via the combination of hypergraph Laplacian and attention matrix of the Transformer. HGraphormer treats hypergraph Laplacian and attention matrix as local and global connections, respectively. The local connections represent the local structure of a hypergraph, which is exploited by existing two-stage message passing techniques. 
The global connection enhances information flow between nodes, even though some nodes have no direct or undirect connection, which further improves the representation power of HGraphormer.  The advantage of incorporating global interactions is experimentally validated in Sec.\ref{sec:gamma}.

In summary, our contributions are fourfold.
\begin{itemize}
\item We propose a new one-stage message passing paradigm for hypergraph node representation learning. It can model global and local information regardless whether there is a direct or indirect connection between nodes. To the best of our knowledge, we are the first work in the field of hypergraphs to use one-stage message passing.
\item We propose HGraphormer, a Transformer-based hypergraph node representation learning framework, which injects structure information by combining the attention matrix and the hypergraph Laplacian.
\item We give theoretical proof that, for hypergraphs, two-stage message passing can be transformed into one-stage message passing. Compared with two-stage, one-stage message passing is more flexible to be extend, which allows the structure of hypergraphs to be more naturally injected into Transformers.
\item Through extensive experiments on five real-world benchmark datasets about node classification task, our HGraphormer technique outperforms strong baseline methods, setting new state-of-the-art performance.
\end{itemize}

\section{Related Work}
This section briefly reviews existing works of hypergraph learning and Transformer on (hyper)graph.
\subsection{Hypergraph Learning}
Learning node representation based on GNNs with graph-based message passing is mature~\cite{GCN2017,GAT2018,GTNs2019,Graphormer2021}. Whereas emulating a graph-based message passing framework for hypergraphs is not straightforward, a hyperedge may consist of more than two nodes, making the interactions among hyperedge complex. HyperSAGE~\cite{arya2020hypersage} uses a two-stage message passing strategy to propagate information through hypergraphs. First, it facilitates neighbourhood node sampling to form hyperedge representations. Then, node representations are aggregated by hyperedges' representations with a power mean function. Similarly, AllSet~\cite{chien2022you} also implements hypergraph neural network layers as compositions of two multiset functions, aggregating node sets and hyperedge sets respectively.
\cite{UniGNN2021} further points out that GNNs can be extended to hypergraphs. Based on the graph-based message passing paradigm of GCN~\cite{GCN2017}, GAT~\cite{GAT2018}, GIN~\cite{GTNs2019} and GraphSAGE~\cite{graphsage2017}, \cite{UniGNN2021} proposes a series of UniGNN (UniGCN, UniGAN, UniGIN, and UniSAGE) following two-stage message passing paradigm. To the best of our knowledge, this series of UniGNN are the state-of-the-art methods for hypergraph node representation learning. 
\subsection{Transformer on (hyper)graph}
Transformer has demonstrated a great power in modelling graph-structured data~\cite{min2022masked,yao2020heterogeneous,georgiev2022heat,chien2022you}. Two typical ways to inject the graph information into the vanilla Transformer are designing positional embedding or attention matrix from graphs. \cite{arya2020hypersage} adopts Laplacian eigenvectors as positional embeddings in Graph Transformer. \cite{hussain2021edge} employs pre-computed SVD vectors of the adjacency matrix as the positional embeddings. While positional embedding is convenient to incorporate graph with Transformer, compressing graph structure into fixed-size vectors (positional embedding) suffers from information loss\cite{graves2016hybrid}. To overcome this issue, another group of works attempts to design attention matrix based on the graph structure. \cite{min2022masked} uses the adjacent matrix to mask the attention matrix. \cite{yao2020heterogeneous} computes the attention matrix based on the extended Levi graph. However, attention module can only capture pairwise relationships rather than high-order relationships of hypergraphs. Recent Transformer-based hypergraph works~\cite{georgiev2022heat,chien2022you} employ Transformer as an encoder module for each stage of message passing. They first encode nodes to form hyperedge representations, then encode hyperedges to produce node representations. Nevertheless, above single-stage encoder strategy does not fully exploit the ability of global modelling of Transformer. 
\section{Preliminaries}
In this section, we introduce the main concepts and formulate the problem. The important notations used in this paper are summarized in Table~\ref{tab:key_notations}.

\subsection{Hypergraph}
A hypergraph is defined as $\mathcal{G}=(\mathcal{V},\mathcal{E})$, where $\mathcal{V}$ is the node set and $\mathcal{E}$ is the hyperedge set. 
A hyperedge $e\in\mathcal{E} = \{v_1,v_2, ...\}$ connects a set of nodes instead of only two in regular graphs. Here each $v_i$ is a node and $\left| e\right| \geq 2 $.
A weighted hypergraph is defined as $\mathcal{G}=(\mathcal{V},\mathcal{E},\mathcal{W})$,  where a positive number $\mathcal{W}_e$,  the weight of hyperedge $e$, is associated with each $e$. A hyperedge $e$ is said to be incident with a node $v$ when $v \in e$. A hypergraph $\mathcal{G}$ can be denoted by an $N\times M$ incidence matrix $\mathbf{H}$, where $N = \left | \mathcal{V}  \right |$ and $M =  \left | \mathcal{E}  \right |$, with entries defined as follows: \begin{equation}\label{eq:attennew}
    \begin{split}
        H_{v,e}=\left
                \{ \begin{array}{ll}
                    1, & \text { if } v \in e \\
                    0, & \text { otherwise} 
                \end{array}
                \right.
    \end{split}
\end{equation}

For a node $v \in \mathcal{V}$ and  a hyperedge  $e \in \mathcal{E}$, their degrees are defined as $d(v)=\sum_{e \in \mathcal{E}} \mathcal{W}_eH_{v, e}$ and $d(e)=\sum_{v \in \mathcal{V}} H_{v, e}$.  Further, $\mathbf{D}_{e}$ and $\mathbf{D}_{v}$ are diagonal matrices denoting the hyperedge degrees and the node degrees, respectively. $\mathbf{W}$ is the diagonal matrix containing the weights of hyperedges. The initial features for all nodes are denoted as $\mathbf{X}$. 

\subsection{Hypergraph Laplacian}\label{sec:Laplacian}
For a given hypergraph $\mathcal{G}$, the incidence matrix $\mathbf{H}$ represents its structure (see Figure~\ref{fig:coauthor} top right). Each row in $\mathbf{H}$ shows a node's connection to each hyperedge, representing high-order relationships. Based on the incidence matrix $\mathbf{H}$, \cite{zhou2006learning} proposes hypergraph Laplacian $\mathbf{L}$ (see Figure~\ref{fig:coauthor} bottom right) where each element represents the connection between nodes. $\mathbf{L}$ is defined as follows,
\begin{equation}\label{eq:laplacian}
\mathbf{L} = \mathbf{D}_{v}^{-1 / 2} \mathbf{H W} \mathbf{D}_{e}^{-1} \mathbf{H}^{\top} \mathbf{D}_{v}^{-1 / 2}
\end{equation}
where $\mathbf{L} \in \mathbb{R}^{N \times N}$. The element $L_{i,k}$ of $\mathbf{L}$ denotes the weight between node $v_i$ and node $v_k$.

\begin{table}[]
\caption{Notations and Definitions}
    \label{tab:key_notations}
	\begin{tabular}{l|l}
		\hline
		Notations      & \multicolumn{1}{c}{Definitions}                                        \\ \hline
		$b,\mathbf{b} ,\mathbf{B} ,\mathcal{B} $& Scalar, vector, matrix and set    \\ \hline
		$\mathcal{G}$  & \begin{tabular}[c]{@{}l@{}}$\mathcal{G}=(\mathcal{V},\mathcal{E})$ indicates a hypergraph. $\mathcal{V}$ are $\mathcal{E}$\\are  node set and hyperedge set, respectively. \\$\mathcal{G}=(\mathcal{V},\mathcal{E},\mathcal{W})$ indicates a weighted hypergraph,\\ a positive number $\mathcal{W}_e$ associated with each \\hyperedge $e$.\end{tabular} \\ \hline
		$\mathcal{E}$   & Hyperedges set of  $\mathcal{G}$    \\ \hline
		$\mathcal{V}$ & Nodes set  of $\mathcal{G}$    \\ \hline
		$\mathcal{W}$ & Hyperedge weights set  of $\mathcal{G}$    \\ \hline
		$v_i,e_j$ &  Node $v_i$ and hyperedge $e_j$ \\ \hline
        $\mathcal{E}_{v_i}$ & \begin{tabular}[c]{@{}l@{}} $\mathcal{E}_{v_i} = \left \{ e \in \mathcal{E} | v_i \in e\right \}$ denotes the incident-edges \\ of node $v_i$. \end{tabular}  \\ \hline
		$M$   & The number of hyperedges on  $\mathcal{G}$,   $M =  \left | \mathcal{E}  \right |$\\ \hline
		$N$ & The number of nodes on  $\mathcal{G}$,  $N = \left | \mathcal{V}  \right |$ \\ \hline
		$\mathbf{H} $   & The incidence Matrix of $\mathcal{G}$ , $\mathbf{H} \in \mathbb{R}^{N \times M}$              \\ \hline
		$\mathbf{L} $   & \begin{tabular}[c]{@{}l@{}} The hypergraph  Laplacian of  $\mathcal{G}$, \\$\mathbf{L} = \mathbf{D}_{v}^{-1 / 2} \mathbf{H W} \mathbf{D}_{e}^{-1} \mathbf{H}^{\top} \mathbf{D}_{v}^{-1 / 2}$, $\mathbf{L} \in \mathbb{R}^{N \times N}$.        \end{tabular}   \\ \hline
		$\mathbf{W},\mathbf{D}_e,\mathbf{D}_v $   &  \begin{tabular}[c]{@{}l@{}}The diagonal matrix of hyperedge weights, \\hyperedge degrees  and  node degrees on  $\mathcal{G}$, \\ respectively.  $\mathbf{W} \in \mathbb{R}^{M \times M}$, $\mathbf{D}_e  \in \mathbb{R}^{M \times M}$\\ and $\mathbf{D}_v  \in \mathbb{R}^{N \times N}$.         \end{tabular}          \\ \hline
            
		$\mathbf{X} $   &  \begin{tabular}[c]{@{}l@{}}Initial features for all nodes on $\mathcal{G}$ , $\mathbf{X} \in \mathbb{R}^{N \times c}$ , \\$c$ is feature dimension.        \end{tabular}        \\ \hline
		$\mathbf{M} $   & \begin{tabular}[c]{@{}l@{}}The attention matrix of Scaled Dot-Product \\Laplacian Attention, $\mathbf{M} \in \mathbb{R}^{N \times N}$        \end{tabular}       \\ \hline
		$\mathcal{N}(v_i)$ & \begin{tabular}[c]{@{}l@{}} Node $v_i$'s \ neighborhood nodes across \\all hyperedges.\end{tabular}     \\ \hline
		$d_h$ & Hidden dimension of HGraphomer  \\ \hline
		$d_k,d_q$ & Dimension of attention Key and attention Value  \\ \hline
		$\mathbf{Z}^l$& The output of $l$th HGraphomer layer. $\mathbf{Z}^l \in \mathbb{R}^{N \times d_h}$\\ \hline
	\end{tabular}
\end{table}

\subsection{Transformer}
The Transformer architecture consists of Transformer layers. The core of each Transformer layer is the self-attention module. Let $\mathbf{Z}^l = [{\mathbf{z}^l_1}^T,...,{\mathbf{z}^l_N}^T] ^T\in \mathbb{R}^{N\times d_h}$ denotes the input of self-attention module of layer $l$, where $d_h$ is the hidden dimension of $\mathbf{Z}^l$ and $\mathbf{z}^l_i \in \mathbb{R}^{1 \times d_h}$ is the hidden representation of layer $l$. The input $\mathbf{Z}^l$ is projected by three matrices  $\mathbf{W}_{Q}  \in \mathbb{R}^{d_h \times d_k}$, $\mathbf{W}_K  \in \mathbb{R}^{d_h \times d_k}$ and $\mathbf{W}_V \in \mathbb{R}^{d_h \times d_q}$ to the corresponding representation matrices $\mathbf{Q,K,V}$. Then the self-attention is  calculated as,
\begin{equation}\label{eq:attention_weight}
    \mathbf{Q} = \mathbf{Z}^l\mathbf{W}_{Q}  ,\quad \mathbf{K} = \mathbf{Z}^l\mathbf{W}_{K}  , \quad \mathbf{V} = \mathbf{Z}^l\mathbf{W}_{V}  ,
\end{equation}
\begin{equation}\label{eq:standard_attention}
    \mathbf{M} = softmax(\frac{\mathbf{Q}\mathbf{K}^T}{\sqrt{d_k}})
\end{equation}
\begin{equation}\label{eq:att}
     Attention(\mathbf{Q},\mathbf{K},\mathbf{V})=\mathbf{M}\mathbf{V}
\end{equation}
We call $\mathbf{M}$ as attention matrix which captures the similarity between queries and keys. As a result, self-attention in Transformer can only model pairwise relationships.  Furthermore, the extension to the multi-head attention is standard and straightforward.

\subsection{Problem Formulation}
\label{def:Problem1} Given a hypergraph $\mathcal{G}=(\mathcal{V},\mathcal{E},\mathcal{W})$, our task is to learn the representations of nodes $v\in\mathcal{V}$ for downstream tasks, such as node classification.

\section{Message Passing in Hypergraph}\label{theory_analysis}
In this section, we first analyse the existing two-stage message passing paradigm on hypergraphs, based on which we propose our first intuitive version of one-stage message passing paradigm for hypergraphs which will be extended to a ``Plus'' version in Sec.~\ref{sec:onestagePlus}.

Most existing HGNNs follow a two-stage message passing paradigm and aggregate node representations and hyperedge representations in stages iteratively. Firstly, node representations are aggregated to generate hyperedge representations through a predefined node aggregation function. Then, a predefined hyperedge aggregation function aggregates hyperedge representations to update node representations. The representation learning formulations for hyperedges and nodes are defined as follows,
\begin{align}
    \mathbf{e}_j &= \phi _1\left (\left\{ \mathbf{v_i} \right\}_{v_i \in e_j}\right )\label{eq:node_aggre1}\\
  \hat{\mathbf{v}_i} &= \phi _2\left ( \left\{ \mathbf{e}_j \right\}_{e_j \in \mathcal{E}_{v_i}}\right ) \label{eq:node_aggre2}
\end{align} 
where $\mathbf{v}_k$ is the node ${v}_k$'s initial representation. $\phi_1$ and $\phi_2$ are permutation-invariant functions for updating hyperedge representation $\mathbf{e}_j$ and node representation $\hat{\mathbf{v}_i}$ from corresponding nodes and hyperedges respectively.  $\mathcal{E}_{v_i} = \left \{ e \in \mathcal{E} | v_i \in e\right \}$ denotes the set of incident hyperedges of node $v_i$. 

\subsection{Two-stage Message Passing}\label{sec:analysis_two} 
\subsubsection{HyperSAGE} \label{sec:HyperSAGE}
As a representative HGNNs-based model, HyperSAGE~\cite{arya2020hypersage} follows the two-stage message passing paradigm. With $\mathcal{N}(v_i)$ denotes node $v_i$'s \ neighborhood nodes across all hyperedges,  HyperSAGE aggregates information with the following formulations,
\begin{align}
    \mathbf{e}_{j} &= (\frac{1}{d_{e_j}}\sum_{v_{k} \in e_j}^{}\mathbf{v}^p_k)^{\frac{1}{p}}\label{eq:hsage1}\\
        {\hat{\mathbf{v}_{i}}}  &= (\frac{1}{d_{v_i}}\sum_{e_j \in \mathcal{E}_i}\frac{d_{e_j}}{\left | \mathcal{N}(v_i) \right |}(\mathbf{e}_j)^p)^{\frac{1}{p}}\label{eq:hsage2}
\end{align}
where $d_{e_j}$ and $d_{v_i}$ are the degree of hyperedge $e_j$ and node $v_i$ respectively; and $p$ is a power value. Following prior work~\cite{arya2020hypersage}, we set $p=1$ as it achieves the best performance. 

This work focuses on the node representation learning task, so we eliminate ${e}_j$ by plugging Eq.~(\ref{eq:hsage1}) into Eq.~(\ref{eq:hsage2}). Finally, Eq.~(\ref{eq:hsage2}) can be rewritten as,
\begin{equation}\label{eq:hsage}
    \begin{split}
        {\hat{\mathbf{v}_{i}}}  =  \frac{1}{d_{v_i}\left | \mathcal{N}(v_i) \right |}\sum_{e_j \in \mathcal{E}_{v_i}}\sum_{v_k \in e_j}^{}\mathbf{v}_k
    \end{split}
\end{equation}

Eq.~(\ref{eq:hsage}) indicates that node representation $\hat{\mathbf{v}_{i}}$  is a weighted sum of representations of all nodes connecting $v_i$ through hyperedges. The weight between $v_k$ and $v_i$ is their co-occurs in hyperedges normalized by $v_{i}$'s degree and the number of its neighbour nodes.

\subsubsection{UniGCN}  \label{sec:UniGCN}
UniGCN~\cite{UniGNN2021} is a recent representative work of two-stage message passsing paradigm. Similar to GCN~\cite{GCN2017}, it propagates message using weighted sum function,
\begin{align}
        \mathbf{e}_j &= \frac{1}{d_{e_j}}\sum _{v_k \in e_j} \mathbf{v}_k\label{eq:unignn1}\\
        \hat{\mathbf{v}_{i}}  &= \frac{1}{\sqrt{d_{v_i}}}\sum _{e_j \in \mathcal{E}_i} \frac{1}{\sqrt{d_{e_j}}} w_{d_{e_j}} \mathbf{e}_j\label{eq:unignn2}
\end{align}
where $w_{d_{e_j}}$ is a specific weight  to the hyperedge degrees. Similar to Eq.~(\ref{eq:hsage}), we merge Eq.~(\ref{eq:unignn1}) and Eq.~(\ref{eq:unignn2}) as follows:
\begin{equation}\label{eq:unignn}
    \begin{split}
        \hat{\mathbf{v}_{i}}  = \frac{1}{\sqrt{d_{v_i}}}\sum _{e_j \in \mathcal{E}_i} \sum _{v_k \in e_j}\frac{w_{e_j}}{{d_{e_j}}^{\frac{3}{2}}} \mathbf{v}_k
    \end{split}
\end{equation}

Similar to Eq.~(\ref{eq:hsage}), the node representation $\hat{\mathbf{v}_{i}}$ in Eq.~(\ref{eq:unignn}) is also a weighted sum of representations of the nodes connected through hyperedges. 

\subsection{One-stage Message Passing}  \label{sec:ext_analysis}
For the two-stage message passing paradigm, messages flow alternatively between nodes and hyperedges (i.e.\ nodes $\rightarrow$ hyperedges $\rightarrow$ nodes $\rightarrow\ldots$), 
in which hyperedges play the role of bridges. Information flows not only start at nodes, but also end at nodes. 
Thus, a natural question arises, \emph{why not model message passing between nodes directly?} Motivated by our derivation in Eq.~(\ref{eq:hsage}) and Eq.~(\ref{eq:unignn}), we propose an intuitive one-stage message passing paradigm to learn node-to-node interactions directly. 

\begin{figure}[h]
    \centering
    \includegraphics{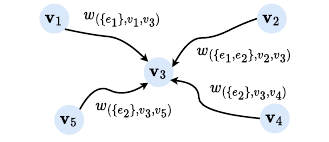} 
    \caption{An example of our proposed one-stage message passing paradigm for hypergraphs, where $\mathbf{v}_i$ represents the features of node $v_i$. $w(\{e_j | e_j \in \mathcal{E}_{v_k,v_i}\},v_k,v_i)$ controling message passing between node $v_i$ and node $v_k$.}
    \label{fig-mpass1}
\end{figure}

We derive one-stage message passing equations from two-stage message passing equations in Section~\ref{sec:analysis_two}, which simplify the message passing flow in hypergraphs. 
Eq.~(\ref{eq:hsage}) and Eq.~(\ref{eq:unignn}) show different functions to learn weights between two connected nodes on incident hyperedges. Hence, an intuitive one-stage message passing paradigm can be formulated as follows,
\begin{equation}\label{eq:one-stage}
    \begin{split}
        \hat{\mathbf{v}_{i}}  = \sum _{v_k \in \mathcal{V}} w(\{e_j | e_j \in \mathcal{E}_{v_k,v_i}\},v_k,v_i)\mathbf{v}_k 
    \end{split}
\end{equation}
where $w(\{e_j | e_j \in \mathcal{E}_{v_k,v_i}\},v_k,v_i)$ is a weight function associated with $v_k$ and  $v_i$ on concurrent hyperedges set of $\mathcal{E}_{v_k,v_i} = \mathcal{E}_{v_k} \bigcap \mathcal{E}_{v_i} $, which controls the amount of message passing between $v_k$ and $v_i$. Eq.~(\ref{eq:one-stage}) can be illustrated through Figure~\ref{fig-mpass1}. 
However, one \textbf{disadvantage} of both the traditional two-stage message passing and the above simple one-stage message passing as shown in Eq.~(\ref{eq:one-stage}) is that there is no message passing between $v_k$ and $v_i$ if $\mathcal{E}_{v_k} \bigcap \mathcal{E}_{v_i} = \varnothing$. In other words, Eq.~(\ref{eq:one-stage}) represents a \emph{local} message passing paradigm, which is not optimal, as in many real-world situations, implicit relationships between nodes may exist even if there is no connection between them. Therefore, messages passing should also be \emph{global} to flow over implicit relationships.

In the next section, we introduce our Transformer-based hypregraph representation method that supports both local and global message passing in the one-stage message passing paradigm.

\section{HGraphormer}
\label{sec:hgraphormer}
In this section, we present our HGraphormer for hypergraph node representation learning. 
We first propose a new message passing paradigm on hypergraphs which is a ``Plus'' version of the basic one-stage message passing paradigm in Sec.~\ref{sec:ext_analysis}. One-stage message passing Plus can model both global and local feature at the same time. Then, we elaborate several vital designs by injecting hypergraph structure into the attention module of Transformer to learn hypergraph node representations, and detail the implementations of HGraphormer.

\subsection{One-stage Message Passing \textit{Plus}}\label{sec:one-stage}
\label{sec:onestagePlus}
Recall Eq.~({\ref{eq:one-stage}}) updates $\hat{\mathbf{v}_{i}}$ with only nodes that are directly connected to node ${v}_i$, which limits the model's ability in capturing global information (e.g.\ unconnected nodes) in hypergraphs. However, it has been demonstrated~\cite{Graphormer2021} that global information is important in graphs. To exploit this global information, we define one-stage message passing ``Plus'' paradigm in Eq.~({\ref{eq:one-stage_glo}}), which enables HGraphormer to capture both global and local information.
\begin{equation}\label{eq:one-stage_glo}
    \begin{split}
        \hat{\mathbf{v}_{i}}  = \sum _{v_k \in \mathcal{V}} (w^{global}_{ki}+w^{local}_{ki})\mathbf{v}_k 
    \end{split}
\end{equation}
where $w^{global}_{ki}$ captures the implicit relationship between $v_k$ and $v_i$. At the same time, $w^{local}_{ki}$ captures the local connections between $v_k$ and $v_i$, which is hypergraph structure. The next subsection will illustrate how to compute $w^{global}$ and $w^{local}$.


\subsection{Structural Injection in HGraphormer}
The self-attention matrix $\mathbf{M}$ defined in Eq.~(\ref{eq:standard_attention}) represents the semantical pairwise correlations between any nodes, which can capture the global information in hypergraphs. This is also the reason why Transformer achieve global receptive field. However, $\mathbf{M}$ does not involve the local relational (structure) information of hypergraph.

Hypergraph Laplacian $\mathbf{L}$, as illustrated in Sec.~\ref{sec:Laplacian}, is the spatial view of hypergraph structure, which highlights the relationships between connected nodes. Hence, we inject hypergraph Laplacian $\mathbf{L}$ to the attention matrix $\mathbf{M}$ in Transformer
The structured attention matrix $\mathbf{A}$ is defined as,
\begin{equation}\label{eq:a_map}
    \mathbf{A} = \gamma \mathbf{M} + (1-\gamma)\mathbf{L}
\end{equation}
where $\gamma$ is a hyper-parameter to balance semantic correlation between nodes and structure of hypergraphs, and we share $\gamma$ and $\mathbf{L}$ across all layers. Here, $M$ and $L$ represent $w^{global}$ and $w^{local}$ in Eq.~\ref{eq:one-stage_glo} respectively. 

\subsection{Design Details}
HGraphormer is built upon the original implementation of classical Transformer encoder~\cite{Vaswani2017}. It consists of \textit{Scaled Dot-Product Laplacian Attention}, \textit{Multi-Head Laplacian Attention} and \textit{HGraphormer Layer}.
\subsubsection{Scaled Dot-Product Laplacian Attention} 
\begin{figure}[h]
	\centering
	\includegraphics[height=160pt]{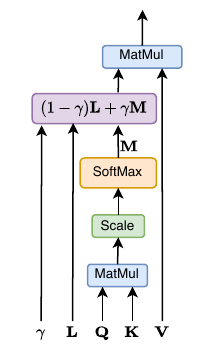} 
	\caption{Scaled Dot-Product Laplacian Attention.}
	\label{fig-att}
\end{figure}

As the core module of HGraphomer, scaled dot-product Laplacian attention encodes the semantic correlations of nodes and the hypergraph structure through one-stage message passing, which is defined as follows,
\begin{equation}\label{eq:ScaledDot-ProductLaplacianAttention}
\begin{split}
    LaplacianAtt(\mathbf{Q},\mathbf{K},\mathbf{V},\mathbf{L},\gamma) &=\mathbf{A}\mathbf{V} \\
    &=\left ( \gamma \mathbf{M} + (1-\gamma)\mathbf{L}\right )\mathbf{V} 
\end{split}
\end{equation}
where $\mathbf{M},\mathbf{Q},\mathbf{K},\mathbf{V}$ can be caculated through  Eq.~(\ref{eq:attention_weight}) and Eq.~(\ref{eq:standard_attention}) with the input $\mathbf{Z}^l$ which is node representations from the $l$-th layer. Figure~\ref{fig-att} is an illustration of this module.

%
\begin{table*}[bht]
    \center
    \caption{Statistic of five real-world hypergraph datasets. 
    $\left | \mathcal{V} \right | $ and $\left | \mathcal{E} \right |$ denotes the number of nodes and edges, respectively.  
    $ avg.\ d_v = \frac{1}{\left |\mathcal{V} \right |}\sum_{v \in \mathcal{V}}d_v$ denotes average node degree.
    $ avg.\ d_e  = \frac{1}{\left |\mathcal{E} \right |}\sum_{e \in \mathcal{E}}d_e$ denotes average hyperedge degree. 
    $avg.\ d_{Neig}  = \frac{1}{\left |\mathcal{V} \right |}\sum_{v \in \mathcal{V}}\left |\mathcal{N}(v_i) \right |$ denotes average neighbourhood node number.
    $c$ the dimension of node features,
    $C$ is the number of classes, and $\eta$ denotes the label rate.}
    \label{tab:statistic}
    \begin{tabular}{@{}cccccc@{}}
        \toprule
        \multirow{2}{*}{Method\quad} & \multicolumn{2}{c}{Co-authorship Data}              & \multicolumn{3}{c}{Co-citation Data}                                                 \\ \cmidrule(l){2-3} \cmidrule(l){4-6} 
                                & \multicolumn{1}{c}{\quad\quad DBLP\quad\quad} & \multicolumn{1}{c}{\quad\quad Cora-ca\quad\quad} & \multicolumn{1}{c}{\quad\quad Pubmed \quad\quad} & \multicolumn{1}{c}{\quad\quad Citeseer \quad\quad} & \multicolumn{1}{c}{\quad\quad Cora-cc\quad\quad } \\ \cmidrule(r){1-6}
        $\left |\mathcal{V} \right |$ & 43,413& 2,708 & 19,717& 3,312 & 2708 \\
        $\left |\mathcal{E} \right |$ & 22,535& 1,072& 7,963& 1,079& 1,579 \\
        $avg.\ d_v$ & 4.7 & 4.2 & 4.3 &  3.2 & 3.0  \\
        $avg.\ d_e$ & 2.41& 1.92 & 9.01 &2.69 & 3.33  \\
        $avg.\ d_{Neig} $  & 21.94 & 13.51 & 65.49 & 6.91  & 6.94  \\
        $C$  & 6& 7& 3&  6& 7 \\
        $c$  & 1,425& 1,433 & 500 &  3,703 & 1,433 \\
        $\eta$  & 4.0\% & 5.2\% & 0.8\% & 4.2\% & 5.2\% \\
        \bottomrule
\end{tabular}
\end{table*}

\subsubsection{Multi-Head Laplacian Attention}
\begin{figure}[h]
	\centering
	\includegraphics[height=160pt]{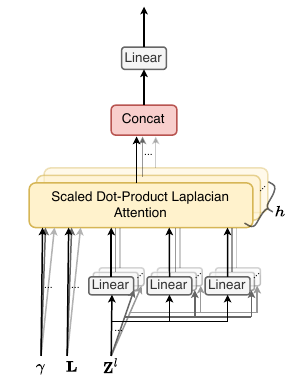} 
	\caption{Multi-Head Laplacian Attention. Multi-Head consists of several attention layers running in parallel.}
	\label{fig-mha}
\end{figure}

Multi-Head Laplacian Attention allows HGraphormer to jointly process information from different representation subspaces. It is defined as follows,
\begin{equation}\label{eq:Mult-head}
    MultiHead(\mathbf{Z}^l,\mathbf{L},\gamma)=Concat(head_1,...,head_h)\mathbf{W}_Z
\end{equation}
where $head_i = LaplacianAtt(\mathbf{Z}^l\mathbf{W}_Q,\mathbf{Z}^l\mathbf{W}_K,\mathbf{Z}^l\mathbf{W}_V,\mathbf{L},\gamma)$. $\mathbf{Z}^l$ is node representations from the $l$-th layer. $\mathbf{W}_Q,\mathbf{W}_K,\mathbf{W}_V$ are linear mapping. $h$ is the number of attention heads, and $\mathbf{W}_Z \in \mathbb{R}^{hd_q \times d_h}$ is a linear function to reduce dimensions. Figure~\ref{fig-mha} illustrates its mechanism.

\subsubsection{HGraphormer Layer}
Unlike the design of the original Transformer encoder, we only apply residual layers and layer normalization (LN) after mutil-head Laplacian attention\footnote{We find the feed-forward blocks here lead to over-fitting.}. We formally characterize the HGraphormer layer as follows,
\begin{equation}\label{eq:HGraphormer}
	\begin{split}
        \mathbf{Z}^{l+1} &= HGraphormer(\mathbf{Z}^l,\mathbf{L},\gamma)) \\
        & =  LN(MultiHead(\mathbf{Z}^l,\mathbf{L},\gamma))+\mathbf{Z}^l
    \end{split}
\end{equation}
\begin{figure}[h]
	\label{fig-hgraphormer}
	\centering
	\includegraphics[width=240pt]{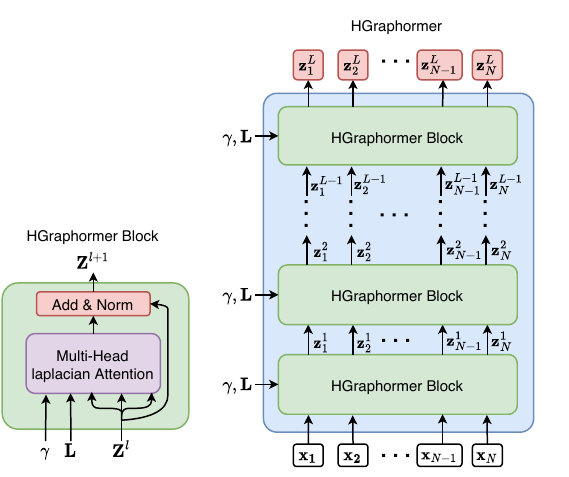} 
	\caption{Framwork of HGraphormer}
\end{figure}
where $\mathbf{Z}^0 = FFD(\mathbf{X}$), and $\mathbf{X} \in \mathbb{R}^{N\times c}$ is the initial feature of nodes. $FFD$ represents feed-forward blocks and $c$ is the dimension of node features. HGraphormer can be built by stacking multiple HGraphormer layers. Figure~\ref{fig-hgraphormer} shows the overall framework of HGraphormer. For the node classification task, the stacked layers are followed by a dense layer and a softmax layer and the loss function is a standard cross-entropy on nodes with ground-truth class labels.

\section{Experiments}
In this section, we conduct experiments on five real-world datasets to answer the following research questions:
\begin{itemize}
    \item (Q1) Does our proposed HGraphormer achieve state-of-art performance in hypergraph node representation learning tasks?
    \item (Q2) Does the "Plus" version of one-stage message passing paradigm outperform that of the \textit{basic} version which models the local information only and how do the global features and local features impact the performance of our HGraphormer framework? 
    \item (Q3) What are the effects of hyperparameters and different modules on the performance?
\end{itemize} 

\subsection{Experimental Setup}

\subsubsection{Datasets.}
We use five widely-used hypergraph datasets~\cite{hypergcn2019,UniGNN2021,cora2008,dblp2015}: DBLP, Cora-ca and Cora-cc from Cora, Pubmed, and CiteSeer. 
Nodes in these hypergraphs represent publications, and hyperedges represent co-authorship/co-citation relationships. 
The co-authorship hypergraphs are built from DBLP
and Cora
, whose hyperedges represent all documents published by the same author. 
The co-citation hypergraphs are constructed from PubMed
, Citeseer
, and Cora, whose hyperedges connect all documents cited by the same documents. 
Since Cora dataset contains both two relationships, we denote Cora as Cora-ca and Cora-cc for co-authorship and co-citation separately. 
These datasets are of varying sizes, with the number of nodes ranges from 2,708 to 43,413, and the number of hyperedges ranges from 1,072 to 22,535, allowing us to evaluate model performance comprehensively.
Other properties also show a great variability amongst the datasets, as can be seen in Table~\ref{tab:statistic}.

\begin{table*}[ht]
\center
\caption{Testing accuracy (\%) of of HGraphormer and other hypergraph learning methods on co-authorship and co-citation datasets for Semi-supervised Hypernode Classification. The best and second best results are bolded and underlined respectively for each dataset. The final row shows the improvement of HGraphormer compared with the second best results.}
    \label{tab:sota}
    \begin{tabular}{@{}lccccc@{}}
        \toprule
        \multirow{2}{*}{Method\quad\quad\quad\quad} & \multicolumn{2}{c}{Co-authorship Data}              & \multicolumn{3}{c}{Co-citation Data}                                                 \\ \cmidrule(l){2-3} \cmidrule(l){4-6} 
                                & \multicolumn{1}{c}{\quad\quad DBLP\quad\quad} & \multicolumn{1}{c}{\quad\quad Cora-ca\quad\quad} & \multicolumn{1}{c}{\quad\quad Pubmed \quad\quad} & \multicolumn{1}{c}{\quad\quad Citeseer \quad\quad} & \multicolumn{1}{c}{\quad\quad Cora-cc\quad\quad } \\ \cmidrule(r){1-6}
        MLP+HLR~\cite{zhou2006learning}     & 63.6 ± 4.7  & 59.8 ± 4.7  & 64.7 ± 3.1  & 56.1 ± 2.6 & 61.0 ± 4.1  \\
        HGNN~\cite{HGNN2019}        & 69.2 ± 5.1  & 63.2 ± 3.1  & 66.8 ± 3.7  & 56.7 ± 3.8 & 70.0 ± 2.9  \\
        FastHyperGCN~\cite{hypergcn2019} & 68.1 ± 9.6  & 61.1 ± 8.2  & 65.7 ± 11.1 & 56.2 ± 8.1 & 61.3 ± 10.3 \\
        HyperGCN~\cite{hypergcn2019}  & 70.9 ± 8.3  & 63.9 ± 7.3  & 68.3 ± 9.5  & 57.3 ± 7.3 & 62.5 ± 9.7  \\
        HyperSAGE~\cite{arya2020hypersage}    & 77.4 ± 3.8  & 72.4 ± 1.6  & 72.9 ± 1.3  & 61.8 ± 2.3 & 69.3 ± 2.7  \\
        UniGAT~\cite{UniGNN2021}      & 88.7 ± 0.2  & 75.0 ± 1.1  & $\underline{74.7}$ ± 1.2  & $\underline{63.8}$ ± 1.6 & 69.2 ± 2.9  \\
        UniGCN~\cite{UniGNN2021}      & $\underline{88.8}$ ± 0.2  & $\underline{75.3}$ ± 1.2  & 74.4 ± 1.0  & 63.6 ± 1.3 & 70.1 ± 1.4  \\
        UniSAGE~\cite{UniGNN2021}      & 88.5 ± 0.2  & 75.1 ± 1.2  & 74.3 ± 1.0  & $\underline{63.8}$ ± 1.3 & $\underline{70.2}$ ± 1.5  \\
        AllSetTransformer~\cite{UniGNN2021}      & 90.25 ± 0.1  & 76.3 ± 1.00 & 85.1 ± 0.22  & 67.3 ± 0.81 & 72.1 ± 0.91  \\
        HGraphormer  & \textbf{92.3} ± 0.19 & \textbf{77.2} ± 1.9 & \textbf{77.2} ± 2.3  & \textbf{67.5} ± 2.6 & \textbf{74.9} ± 2.8  \\ \midrule
          Improvement  & 4.29\%$\uparrow$  & 2.52\%$\uparrow$  & 3.35\%$\uparrow$  & 5.80\%$\uparrow$ & 6.70\%$\uparrow$   \\\bottomrule
    \end{tabular}
\end{table*}

\begin{figure*}[h]
  \centering
  \includegraphics[width=\linewidth]{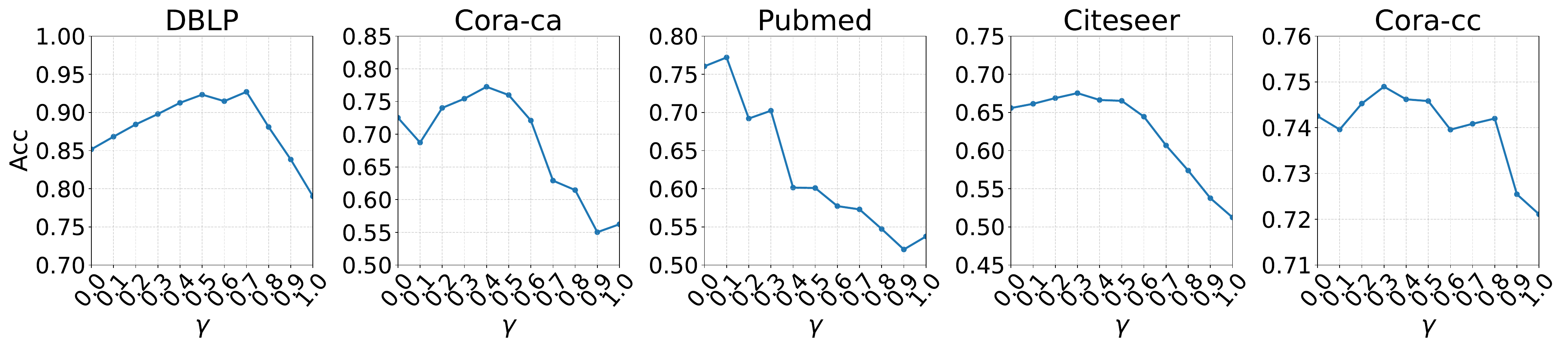} 
  \caption{Test accuracy by varying $\gamma$ in $\gamma \mathbf{M} + (1-\gamma)\mathbf{L}$ (Eq.~\ref{eq:a_map}) on five datasets, where $Acc$ is accuracy}
  \label{fig:gamma}
\end{figure*}
\subsubsection{Task and Evaluation Protocols.} 
We use node classification to evaluate the effectiveness of our representation learning model following the existing hypergraph node representation learning works~\cite{HGNN2019,HGNNPlus2022,dhgnn2019}. More specifically, the task is to predict the topic (e.g., CS and Math) for each node (multi-class classification). Node class label rate $\eta$ on five datasets ranges from 0.8\% to 5.2\%, which means that essentially our task here is semi-supervised node classification task.
We adopt the 10-fold cross validation in previous works~\cite{hypergcn2019,UniGNN2021} for a fair and standard evaluation. More specifically, in each split, for each topic, the labels of 10\% of the nodes belonging to that topic are hidden. Evaluation is performed on this set of 10\% of nodes without labels. 
This process in repeated 10 times to obtain the mean and standard deviation of classification accuracy for each model. 
\subsubsection{Baseline Methods.}
We compare HGraphormer method with eight strong baseline models, which are Multi-Layer Perceptron with explicit Hypergraph Laplacian Regularization (MLP+HLR)~\cite{zhou2006learning}, HyperGraph Neural Network (HGNN)~\cite{HGNN2019}, Fast HyperGraph Convolutional Network (FastHyperGCN)~\cite{hypergcn2019}, HyperGraph Convolutional Network (HyperGCN), HyperSAGE~\cite{arya2020hypersage}, UniGAT, UniGCN and UniSAGE~\cite{UniGNN2021} with full batch. 
On all datasets, the feature of each node is initialised with the bag-of-words representation of documents. We adopt the settings of all baseline models from~\cite{srivastava2014dropout}, which are the optimal settings for these models.
\subsubsection{Parameter Settings.}
We use the Adam optimizer with a learning rate of 0.01 and the weight decay of 0.0005. Activation function is PReLU~\cite{PReLU}. 
To prevent overfitting, we add a dropout layer after each HGraphormer layer output. $d_k, d_q$ are tuned in the range of \{16, 32, 64, 128\}.


\subsection{Overall Results (Q1)}
We implemented HGraphormer for the downstream task of semi-supervised node classification on hypergraphs to compare with state-of-the-art methods.  
Table ~\ref{tab:sota} summarizes the mean and standard deviation of classification accuracy of all the compared methods on the five datasets. We present the major findings below. 

\begin{itemize}
\item HGraphormer consistently outperforms all the baselines with a noticeable performance lift, achieving \textit{a new state-of-the-art} in the node classification task on hypergraphs. Compared with all the second best methods over the five datasets, the average improvement of HGraphormer is 4.53\%. This considerable performance improvement demonstrates the effectiveness of the overall design of HGraphormer.
\item HGraphormer gains more significant improvements on the datasets Cora-cc and Citeseer compared with baselines, with increases of 6.70\% and 5.80\% over the second best models respectively, while the average improvement is 4.53\% across all five datasets. The possible reason is that by modelling the global information, HGraphormer is able to better mitigate the node connection sparsity problem as the node connections on these two datasets are sparser compared with other datasets. The average number of neighbor nodes is usually used to measure the connectivity between nodes in (hyper)graphs. As shown in Table~\ref{tab:statistic}, $avg.\ d_{Neig}$, which denotes the average number of neighbourhood nodes, is 6.91 and 6.94 on Citeseer and Cora-cc respectively, which are significantly lower than 21.94, 13.51 and 65.49 of the other three datasets. 
\item MLP+HLR~\cite{zhou2006learning} performs consistently the worst across all datasets. The possible reason is that all the other methods are based on the message passing scheme except MLP+HLR. This indicates the strength of message passing on hypergraph representation learning.
\item UniGAT, UniGCN and UniSAGE achieve the second-best performance on different datasets.  This shows that, apart from our proposed HGraphormer, there is no best baseline which performs consistently best across all datasets. This further validates the superiority of HGraphormer which is able to produce consistent best performance on all datasets. 
\end{itemize}


\begin{figure*}[th]
  \centering
  \includegraphics[width = \linewidth]{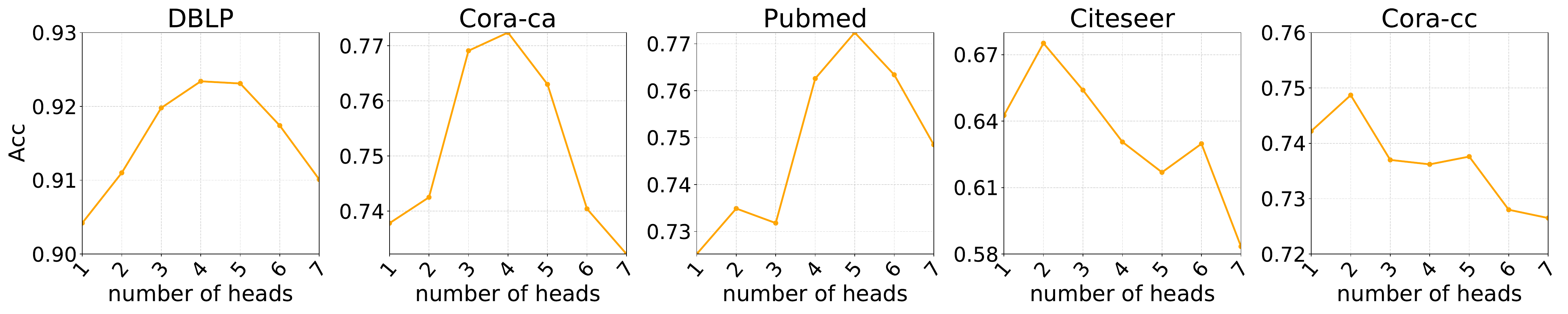} 
  \caption{Accuracy influenced by different heads settings}
  \label{fig:heads}
\end{figure*}

\subsection{Effect of Global and Local Features (Q2)}\label{sec:gamma}
A core contrlibution of HGraphomer is the injection of structure information into Transformer through $\gamma \mathbf{M} + (1-\gamma)\mathbf{L}$, which incorporates the structure of the hypergraph and semantic correlations between nodes.
To answer Q2, we vary the value of the hyperparameter $\gamma$ as it controls the balance between global and local message passing. When $\gamma=0$, only local information is modelled as only the Laplacian is fed to the model; while when $\gamma=1$, only global information is modelled as the hypergraph Laplacian, which captures the structure information, is completely disgarded from the model.
For each dataset, we tune hyperparameter $\gamma$ from 0 to 1 with the step size of 0.1, i.e.\ $\gamma\in\{0, 0.1, 0.2, \ldots, 0.9, 1.0\}$. In Figure~\ref{fig:gamma}, we observe a consistent trend among all datasets, and elaborate our findings below. 
\begin{itemize}
\item HGraphomer achieves optimal performance when both local and global information are taken into account, i.e., $\gamma$ is neither 0 or 1. 
This shows that both local information and global information are essential to a model's performance, which validates the motivation and effectiveness of our design of one-stage message passing \textit{plus} compared with its basic version that only models local information. 
\item On all five datasets, the results when $\gamma=0$ are consistently better than when $\gamma=1$. This shows that compared with global module $\mathbf{M}$, the local structural module $\mathbf{L}$ has a higher impact on the performance of HGraphomer. The direct connection between nodes is the main path of information transmission. 
\item The global structural module $\mathbf{M}$ based on semantic correlation plays a complementary role of local information in the one-stage message passing \textit{plus} paradigm. From $\gamma = 0$ to $\gamma = \gamma_{optimal}$, the accuracy rising slowly with the increase of $\gamma$. The accuracy drops off rapidly as $\gamma$ increases beyond $\gamma_{optimal}$ to 1. 
This observation demonstrates the importance of incorporating global information, while making sure local information is sufficiently represented. 
\item Global information (semantic correlations) can achieve competitive results in the scenarios where local information (real structure) cannot be obtained. 
When $\gamma = 1$, only global information is used.
However, HGraphomer still outperforms HyperSAGE on the DBLP datast (0.77) and UniSAGE on the Cora-cc dataset (0.72) with the accuracy of 0.77 and 0.74 respectively. 
\end{itemize}

\subsection{Parameter Sensitivity Study (Q3)}
We further investigate the parameters and modules of HGraphomer in order to provide insights on how to best optimize our framework for node representation learning.
\subsubsection{Residual Layer} 
Table~\ref{tab:layers} summarizes the accuracy influenced by different residual and layer settings. We observe a consistent trend in all datasets that HGraphomer with residual shows considerable improvement than it without residual, especially when $L > 1$. And before $L$ reaches a certain critical value, the accuracy rise with the increase of $L$, because residual layer makes HGraphformer deeper and deep networks allow information to travel further along the hypergraph structure. 

\begin{table}[]
\center
\caption{Accuracy influenced by different residual and layer settings. Best result is marked bold. $L$ denotes the number of HGraphormer layers. $R$ denotes residual. W/O and W/ denote "Without" and "With" residual.}
\label{tab:layers}
\begin{tabular}{ccccccc}
\toprule
$R$     & $L$ & DBLP   & Cora-ca & Pubmed & Citeseer & Cora-cc \\ \midrule
\multirow{7}{*}{W/O} & 1            & 0.9013 & 0.7583  & 0.7606 & 0.6666   & 0.7473  \\
                     & 2            & 0.8802 & 0.7204  & 0.7551 & 0.3589   & 0.7407  \\ 
                     & 3            & 0.5732 & 0.3878  & 0.4569 & 0.2636   & 0.7038  \\
                     & 4            & 0.2863 & 0.3287  & 0.4515 & 0.2490    & 0.2624  \\
                     & 5            & 0.2632 & 0.3156  & 0.4454 & 0.2493   & 0.2581  \\
                     & 6            & 0.2513 & 0.3131  & 0.4453 & 0.2448   & 0.2393  \\
                     & 7            & 0.2354 & 0.3044  & 0.4409 & 0.2571   & 0.1982  \\ \midrule
\multirow{7}{*}{W/}  & 1            & 0.9220  & 0.7550   & 0.7495 & 0.6320    & 0.7474  \\
                     & 2            & \textbf{0.9231} & 0.7652  & 0.7576 & 0.6484   & \textbf{0.7493}  \\
                     & 3            & 0.9099 & \textbf{0.7722}  & 0.7587 & 0.6620    & 0.7408  \\ 
                     & 4            & 0.8824 & 0.4943  & 0.7645 & 0.6641   & 0.7435  \\
                     & 5            & 0.8704 & 0.3605  & \textbf{0.7722} & \textbf{0.6755}   & 0.7402  \\
                     & 6            & 0.8579 & 0.3250   & 0.7685 & 0.6593   & 0.7340   \\ 
                     & 7            & 0.8079 & 0.3095  & 0.7442 & 0.6435   & 0.7291  \\ \bottomrule
\end{tabular}
\end{table}

\subsubsection{Heads}
 Figure~\ref{fig:heads} shows the accuracy influenced by different attention head settings. We observe that when HGraphormer achieves optimal, the number of head is $> 1$. And it happened cross all five datesets. This observation confirms Multi-Head Laplacian Attention allows HGraphormer to jointly process information from different representation subspaces.

%




\section{CONCLUSION AND FUTURE WORK}
In this work, we have designed a one-stage message passing paradigm to model both global and local information for node representation learning on hypergraphs. 
Under this paradigm, we propose HGraphormer, a novel framework that injects hypergraph structure into Transformer, which consistently performs better than recent state-of-the-art methods on five real-world datasets.
Our future work will extend HGraphormer for broader downstream tasks of hypergraphs, such as hyperedge prediction and hypergraph classification.

\printcredits

\bibliographystyle{cas-model2-names}
\bibliography{main}
%



\end{document}